\def\year{2020}\relax
\documentclass[letterpaper]{article} 
\usepackage{aaai20}  
\usepackage{times}  
\usepackage{helvet} 
\usepackage{courier}  
\usepackage[hyphens]{url}  
\usepackage{graphicx} 
\usepackage{graphicx}  
\usepackage{comment}

\usepackage{lipsum, multicol}
\usepackage{epsfig}
\usepackage{amssymb}
\usepackage{adjustbox}
\usepackage{xcolor}
\usepackage{amsmath}
\usepackage{float}
\usepackage[utf8]{inputenc} 
\usepackage[T1]{fontenc}    
\usepackage{url}            
\usepackage{booktabs}       
\usepackage{amsfonts}       
\usepackage{nicefrac}       
\usepackage{microtype}      
\usepackage{algorithm}
\usepackage{graphicx} 
\usepackage{caption}
\usepackage{subfig}
\usepackage{algorithmic}
\usepackage{booktabs}
\usepackage{multirow}
\usepackage[hyphenbreaks]{breakurl}
\usepackage[hyphens]{url}
\usepackage{adjustbox}
\usepackage{wrapfig}

\newcommand{\eat}[1]{}

\RequirePackage{latexsym}
\RequirePackage{amsmath}
\RequirePackage{amssymb}
\RequirePackage{bm}

\definecolor{magenta}{cmyk}{0,1,0,0}
\definecolor{mygreen}{rgb}{.1,1,.1}
\newcommand{\Magenta}[1]{\color{magenta}{#1}\color{black}}

\ifx\note\undefined
\newcommand{\note}[1]{{\bf \Magenta{#1}}}
\else
\renewcommand{\note}[1]{{\bf \Magenta{#1}}}
\fi

\DeclareMathOperator{\hb}{\mathbf{h}}

\DeclareMathOperator{\vb}{\mathbf{v}}

\DeclareMathOperator{\xb}{\mathbf{x}}
 \DeclareMathOperator{\yb}{\mathbf{y}}

\DeclareMathOperator{\Xb}{\mathbf{X}}

\newcommand{\thetab}{{\bm{\theta}}}
\newcommand{\xib}{{\bm{\xi}}}

\ifx\BlackBox\undefined
\newcommand{\BlackBox}{\rule{1.5ex}{1.5ex}}  
\fi

\ifx\proof\undefined
\newenvironment{proof}{\par\noindent{\bf Proof\ }}{\hfill\BlackBox\\[2mm]}
\fi

\ifx\theorem\undefined
\newtheorem{theorem}{Theorem}
\fi
\ifx\example\undefined
\newtheorem{example}{Example}
\fi
\ifx\lemma\undefined
\newtheorem{lemma}[theorem]{Lemma}
\fi

\ifx\corollary\undefined
\newtheorem{corollary}[theorem]{Corollary}
\fi

\ifx\definition\undefined
\newtheorem{definition}[theorem]{Definition}
\fi

\ifx\proposition\undefined
\newtheorem{proposition}[theorem]{Proposition}
\fi

\ifx\remark\undefined
\newtheorem{remark}[theorem]{Remark}
\fi

\ifx\conjecture\undefined
\newtheorem{conjecture}[theorem]{Conjecture}
\fi

\ifx\factoid\undefined
\newtheorem{factoid}[theorem]{Fact}
\fi

\ifx\axiom\undefined
\newtheorem{axiom}[theorem]{Axiom}
\fi

\newcommand{\phib}{\pmb{\phi}}
\newcommand{\psib}{\pmb{\psi}}

\mathchardef\Theta="7102 \mathchardef\theta="7112

\ifx\br\undefined
\newcommand{\br}{\color{red}}
\else
\renewcommand{\br}{\color{red}}
\fi

\definecolor{lila}{rgb}{0.2,0,0.5}
\definecolor{rot}{rgb}{0.8,0,0}
\definecolor{blau}{rgb}{0,0,0.5}
\definecolor{gruen}{rgb}{0,0.5,0}
\definecolor{grau}{rgb}{0.85,0.8,0.85}
\definecolor{mygreen}{rgb}{.1,1,.1}

\ifx\bl\undefined
\newcommand{\bl}{\hspace*{0.8cm}}
\else
\renewcommand{\bl}{\hspace*{0.8cm}}
\fi

\def\LSTM{\textsf{LSTM}}

\def\Dec{\textsf{Dec}}
\def\MMD{\textsf{MMD}}

\ifx\remark\undefined
\newtheorem{remark}{Remark}
\fi

\definecolor{ublue}{HTML}{00274c}
\definecolor{uyellow}{HTML}{ffcb05}
\definecolor{yangblue}{HTML}{40658f}
\definecolor{yangblood}{HTML}{de1738}

\title{Learning Diverse Stochastic Human-Action Generators by \\Learning Smooth Latent Transitions }
\author{Zhenyi Wang\textsuperscript{\rm 1}, Ping Yu\textsuperscript{\rm 1}, Yang Zhao\textsuperscript{\rm 1}, Ruiyi Zhang\textsuperscript{\rm 2}, Yufan Zhou\textsuperscript{\rm 1}, Junsong Yuan\textsuperscript{\rm 1}, Changyou Chen\textsuperscript{\rm 1} \\
\textsuperscript{\rm 1} State University of New York at Buffalo\\
\textsuperscript{\rm 2} Duke University \\
\textsuperscript{\rm 1} \{zhenyiwa, pingyu, yzhao63, yufanzho, jsyuan, changyou\}@buffalo.edu\\
\textsuperscript{\rm 2} ryzhang@cs.duke.edu\\
}

\begin{document}

\maketitle

\begin{abstract}
 Human-motion generation is a long-standing challenging task due to the requirement of accurately modeling complex and diverse dynamic patterns. Most existing methods adopt sequence models such as RNN to directly model transitions in the original action space. Due to high dimensionality and potential noise, such modeling of action transitions is particularly challenging. In this paper, we focus on skeleton-based action generation and propose to model smooth and diverse transitions on a latent space of action sequences with much lower dimensionality. Conditioned on a latent sequence, actions are generated by a frame-wise decoder shared by all latent action-poses. Specifically, an implicit RNN is defined to model smooth latent sequences, whose randomness (diversity) is controlled by noise from the input. Different from standard action-prediction methods, our model can generate action sequences from pure noise without any conditional action poses. Remarkably, it can also generate unseen actions from mixed classes during training. Our model is learned with a bi-directional generative-adversarial-net framework, which not only can generate diverse action sequences of a particular class or mix classes, but also learns to classify action sequences within the same model. Experimental results show the superiority of our method in both diverse action-sequence generation and classification, relative to existing methods.
 
\end{abstract}

\section{Introduction}

Human-action generation is an important task for modeling dynamic behavior of human activities, with vast real applications such as video synthesis \cite{wang2018vid2vid}, action classification \cite{interpre2017,repactrec2017,stgcn2018,spatio2016,Shahroudy_2016_CVPR,hrnn2015} and action prediction \cite{julieta2017motion,pred_eccv2018,hpgan2017}. Directly generating human actions from scratch is particularly challenging due to the complexity and high-dimensionality of natural scenes. One promising workaround is to first generate easier-to-deal-with skeleton-based action sequences, based on which natural sequence are then rendered. This paper thus focuses on skeleton-based action-sequence generation. 

\begin{figure}[t!]
    \centering
    \includegraphics[width=\linewidth]{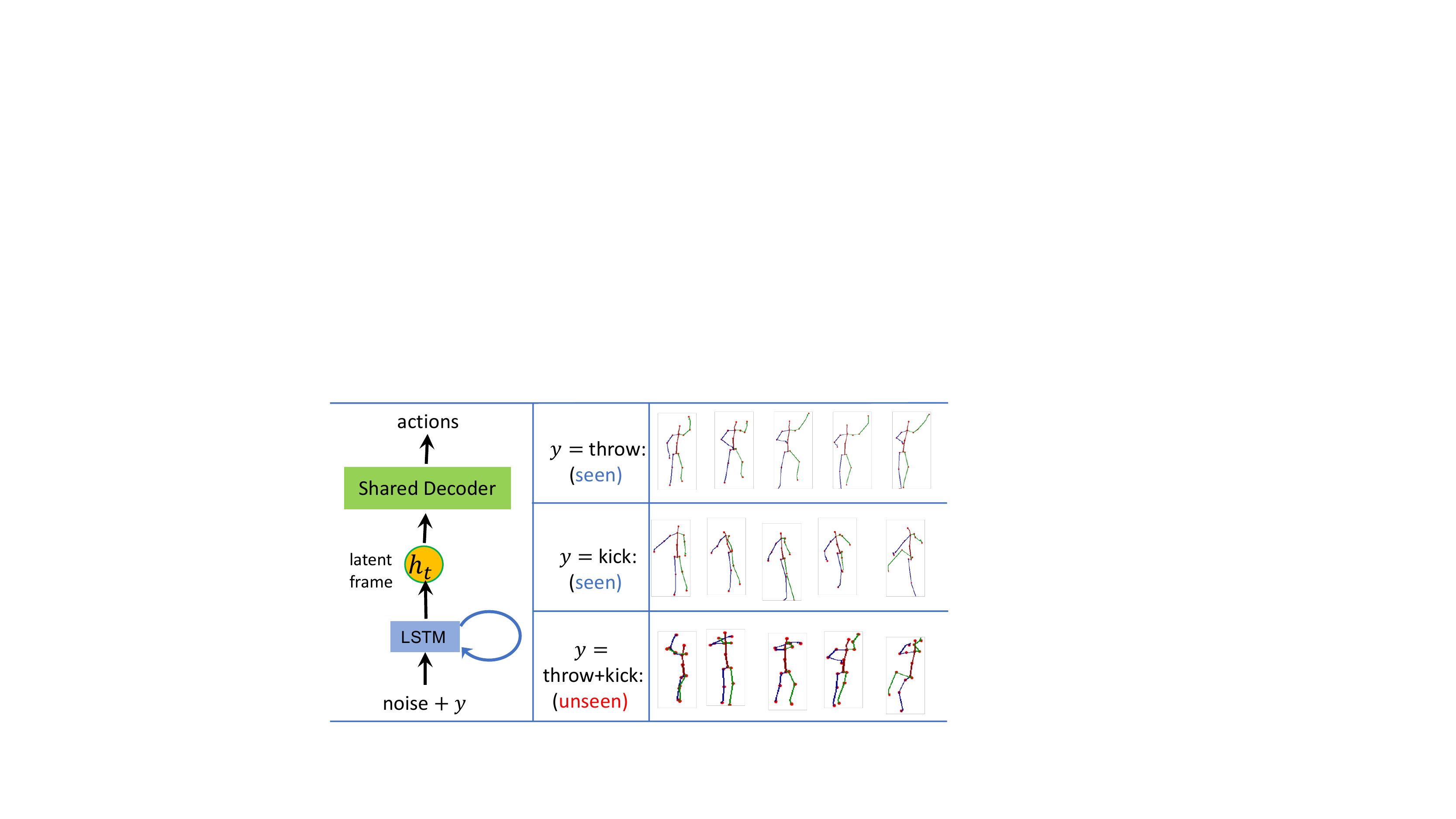}
    \caption{Generating sentences from pure noise, our model can learn smooth latent-frame transitions via an {\em implicit LSTM-based} RNN, which are then decoded to an action sequence via a shared decoder. The action sequence endows a flexible implicit distribution induced by the input noise. Our model not only can generate actions whose class are seen during training ({\it e.g.}, throw, kick), but also can generate actions of unseen mixed classes ({\it e.g.}, throw+kick).}
    \label{fig:model}
    \vspace{-0.5cm}
\end{figure}

Skeleton-based human action generation can be categorized into action synthesis (also referred to generation) \cite{Kovar2002} and prediction. Action synthesis refers to synthesizing a whole action sequence from scratch, with controllable label information; whereas action prediction refers to predicting remaining action-poses given a portion of seed frames. These two tasks are closely related, {\it e.g.}, the latter can be considered as a conditional variant of the former. In general, however, action synthesis is considered more challenging due to little input information available. Existing action-prediction methods can be categorized into deterministic \cite{julieta2017motion,rep2017,pred_eccv2018,RTN2018} and stochastic   \cite{hpgan2017,BIHMP-GAN2019,VAEmotion2017}  approaches. Predicted action sequences in deterministic approaches are not associated with randomness; thus, there is no variance once input sub-sequences are given. By contrast, stochastic approaches can induce probability distributions over predicted sequences. In most cases, stochastic (probabilistic) approaches are preferable as they allow one to generate different action sequences conditioned on the same context. 

For diverse action generation, models are required to be stochastic so that the synthesis process can be considered as drawing samples from an action-sequence probability spaces. As a result, one approach for action synthesis is to learn a stochastic generative model, which induces probability distributions over the space of action sequences from which we can easily sample. Once such a model is learned, new actions can be generated by merely sampling from the generative model.

Among various deep generative models, the generative adversarial network (GAN) \cite{NIPS2014_gan} is one of the state-of-the-art methods, with applications on various tasks such as image generation \cite{ma2017pose}, characters creation \cite{Jin2017}, video generation \cite{Vondrick2016} and prediction \cite{repvideo2017,alex2018,dual2017}. However, most existing GAN-based methods for action generation consider directly learning frame transitions on the original action space. In other words, these works define action generators with recurrent neural networks (RNNs) that directly produce action sequences \cite{dual2017,VAEmotion2017,alex2018,RNN-GAN2018}. However, these models are usually difficult to train due to the complexity and high-dimensionality of the action space.

In this paper, we overcome this issue by breaking the generator into two components: a smooth-latent-transition component (SLTC) and a global skeleton-decoder component (GSDC). Figure~\ref{fig:model} illustrates the key features of and some results from our model. Specifically,
\begin{itemize}
    \item The SLTC is responsible for generating smooth latent frames, each of which corresponds to the latent representation of a generated action-pose. The SLTC is modeled by an {\em implicit LSTM}, which takes a sequence of independent noise plus a one-hot class vector as input, and outputs a latent-frame sequence. Our method inherits the advantage of RNNs, which could generate diverse length sequences but on a much lower-dimensional latent space, an advantage over existing methods such as \cite{Cai_2018_ECCV}.
    \item The GSDC is responsible for decoding each latent frame to an output action pose, via a shared (global) decoder implemented by a deep neural network (DNN). Note that at this stage, only a mapping from a single latent frame to an action-pose needs to be learned, {\it i.e.}, no sequence modeling is needed, making generation much simpler.
\end{itemize}

Our model is learned by adopting the bi-directional GAN framework \cite{ali2017,afl2017}, consisting of a stochastic action generator, an action classifier and an action discriminator. These three networks compete with each other adversarially. At equilibrium, the generator can learn to generate diverse action sequences that match the training data. In addition, the classifier is able to learn to classify both real and synthesized action sequences. Our contributions are summarized as follows:

\begin{itemize}
    \item We propose a novel stochastic action sequence generator architecture, which benefits from an ability to learn smooth latent transitions. The proposed architecture eases the training of the RNN-based generator for sequence generation, and at the same time can learn to generate much higher-quality and diverse actions.
    \item We propose to learn an action-sequence classifier simultaneously within the bi-directional GAN framework, achieving both action generation and classification.
    \item Extensive experiments are conducted, demonstrating the superiority of our proposed framework.
\end{itemize}

\section{Related Works}
Skeleton-based action prediction has been studied for years. One of the most popular methods for human motion prediction (conditioned on a portion of seed action-poses) is based on recurrent neural networks \cite{julieta2017motion,pred_eccv2018,BIHMP-GAN2019}.
For skeleton-based action generation, switching linear models \cite{Vladimir2001,contactdyna2005,learninf2005} were proposed to model stochastic dynamics of human motions. However, it is difficult to select a suitable number of switching states for best modeling. Furthermore, it usually requires a large amount of training data due to the large model size. Restricted Boltzmann Machine (RBM) also has been applied for motion generation \cite{pos_nips2007,temporal2008,cboltz2009}. However, inference for RBM is known to be particularly challenging. Gaussian-process latent variable models \cite{Jack08,Urtasun2008} and its variants \cite{multifact2007} have been applied for this task. One problem with such methods, however, is that they are not scalable enough to deal with large-scale data.

For deep-learning-based methods, RNNs are probably one of the most successful models \cite{rnndynamics2015}. However, most existing models assume output distributions as Gaussian or Gaussian mixture. Different from our {\em implicit representation}, these methods are not expressive enough to capture the diversity of human actions.


In contrast to action prediction, limited work has been done for diverse action generation, apart from some preliminary work. Specifically, the motion graph approach \cite{motiongraph2012} needs to extract motion primitives from prerecorded data; the diversity and quality of action will be restricted by way of defining the primitives and transitions between the primitives. Variational autoencoder and GAN have also been applied in \cite{VAEmotion2017,RNN-GAN2018,gan_act2018} for motion generation. However, these methods directly learn motion transitions with an RNN, and the error of current frame will be accumulate into the next frame, thus making it inapplicable to generate long action sequences, especially for aperiodic motions such as eating and drinking. 

Another distinction between our model and existing methods is that the latter typically require some seed action-frames as input to a generator \cite{visualdynamics16,hpgan2017,RNN-GAN2018,gan_act2018}, which is learned based on the GAN framework; whereas our model is designed to generate action sequence from scratch, and learned based on the bi-direction GAN framework in order to achieve simultaneous action generation and classification. 

\section{The Proposed Model}

\begin{figure}[t!]
	\centering
    \includegraphics[width=1.0\linewidth]{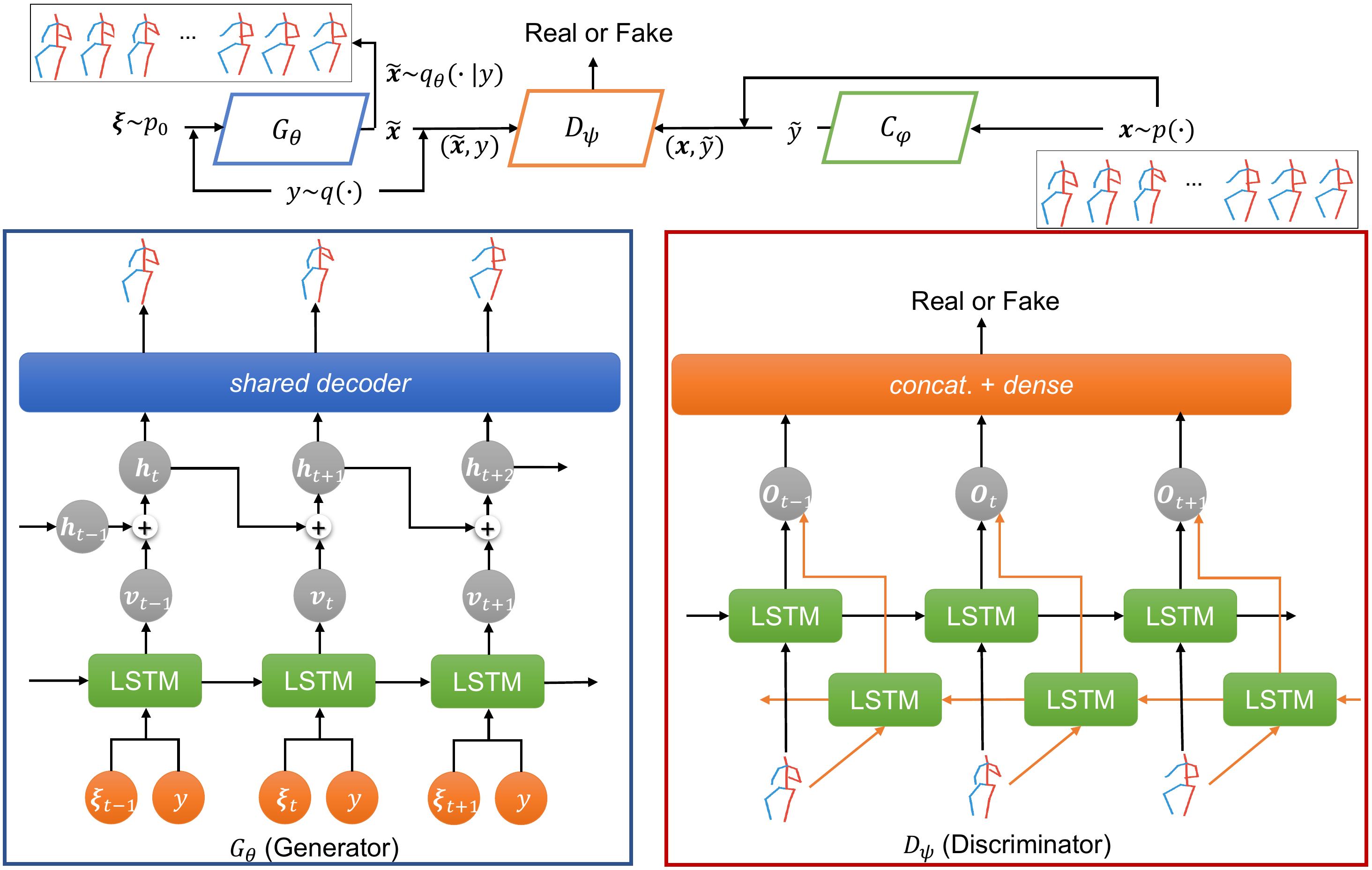}
	\caption{The proposed action-generation model (top), with detailed structures of action sequence generator ($G_{\thetab}$), discriminator ($D_{\psib}$). The classifier ($C_{\phib}$) is the same as $D_{\psib}$ except that it outputs a class label instead of a binary value.} 
	\label{fig:gan_struc}
\end{figure}

We first illustrate our whole model in Figure~\ref{fig:gan_struc}, followed by detailed descriptions of specific components.

\subsection{Problem Setup and Challenges}

Our training dataset is represented as $\Xb \triangleq \{(\xb_1^{(i)}, \cdots, \xb_{T_i}^{(i)}, \yb^{(i)})\}_i$, where $\xb_t^{(i)}\in \mathbb{R}^d$ represents one action-pose of dimension $d$; $T_i$ is the length of the sequence; and $\yb^{(i)}$ is the corresponding one-hot label vector of the sequence\footnote{Our model can also be applied to data without labels by simply removing $\yb^{(i)}$ from the generator. We focus on the one with labels.}. Our basic goal is to train a stochastic sequence generator $G_{\thetab}$ using a DNN parametrized by $\thetab$. Hence, given a label $\yb$ and a sequence of random noises $\xib$ (specified latter), $G_{\thetab}$ is supposed to generate a new action sequence following
\begin{align}\label{eq:generator}
    (\tilde{\xb}_1, \cdots, \tilde{\xb}_{T}) = G_{\thetab}(\yb, \xib)~,
\end{align}
where $T$ is the length of the sequence that can be specified flexibly in $G_{\thetab}$.

\begin{remark}
Similar to implicit generative models such as GAN, we call the generator form \eqref{eq:generator} an {\em implicit generator}, in the sense that the generated sequence $(\tilde{\xb}_1, \cdots, \tilde{\xb}_T)$ is a random sequence endowing an implicit probability distribution with an unknown density function induced by the random noise $\xi$. A traditional way of modeling action sequences usually defines $G_{\thetab}$ as an RNN, which typically defines a Gaussian distribution (explicit) for $\tilde{\xb}_t$, referred to as explicit modeling. An implicit distribution is typically much more flexible than explicit distributions as the density is not restricted to a particular distribution class.
\end{remark}

\paragraph{Challenges}
There are a few challenges. The first one relates to how to define an expressive-enough generator for diverse action generation. We adopt an implicit model of action sequences without explicit form assumption; Thus, the generator benefits from better representation power to generate more sophisticated, higher-quality and diverse action sequences. 
The second challenge is to find an appropriate generator structure. One straightforward way is to define $G_{\thetab}$ as an RNN that outputs action sequences directly, similar to \cite{VAEmotion2017,RNN-GAN2018}. However, it is well known that an RNN with high-dimensional outputs is challenging to train \cite{diffrnn2013}. In recent years, attention \cite{Bahdanau2015} and the Transformer model \cite{Vaswani2017} have been developed to enhance/replace RNN-based models. Attention and Transformer are used for addressing the long-term dependency problem in seq2seq-based models. They are not directly applicable to our setting because our model is not a simple seq2seq model, as our inputs are purely random noise. To this end, we propose a novel generator structure, where smooth \textit{latent-action transitions} are first inferred via an RNN, which are then fed into a shared frame-wise decoder (non-sequential) to map all latent poses to their corresponding action-poses. The detailed structure of the generator is illustrated in Figure~\ref{fig:gan_struc} and described below.

\subsection{Stochastic Action-Sequence Generator}
Our action-sequence generator consists of two components, SLTC and GSDC. The detailed structure of the generator is illustrated as $G_{\thetab}$ in Figure~\ref{fig:gan_struc}.

\paragraph{Learning smooth latent transitions}
Instead of directly modeling sequential transitions in the action space, we propose to model them in a latent action-sequence space. To this end, we decompose $G_{\thetab}$ as compositions of an {\em implicit LSTM} and a shared frame-wise decoder. The LSTM generator (a.k.a.\! SLTC) models smooth latent action transitions, and the shared decoder (a.k.a.\! GSDC) models frame-wise mapping from latent space to action space. Specifically, denote $\hb_t$ to be the latent representation of an action-pose $\xb_t$. We define $(\hb_1, \cdots, \hb_T)$ to be outputs of an {\em implicit LSTM}, written as:
\begin{align}\label{eq:sltc}
    (\hb_1, \cdots, \hb_T) = \LSTM\left(\xib_1, \cdots, \xib_T, \yb; \thetab_1\right)~,
\end{align}
where $(\xib_t, \yb)$ is the input of the LSTM at time $t$ (the noise $\xib_t$'s are independent of each other for all $t$); and $\thetab_1 \subset \thetab$ is the parameter of the LSTM network. We called \eqref{eq:sltc} {\em implicit LSTM} because its input consists of independent noise $\xib_t$ at each time in both training and testing (generation) stages (please see the generator graph in Figure~\ref{fig:gan_struc}). This generator is different from standard LSTM where the output of previous time will be used as input of current time in the testing stage. In addition, the noise in the input would induce a much more flexible implicit distribution on $\hb_t$; whereas standard LSTM defines an explicit distribution such as Gaussian, restricting the representation power. Another advantage of adopting an implicit LSTM as a latent-frame generator is that the length of a generated action sequence could be induced from the latent space instead of the action space. In general, the dimension of $\hb_t$ is much smaller than action poses, making the training of the LSTM easier. Finally, modeling latent representations with an LSTM also allows latent transitions to be smooth, that is the desired property of action sequence generation. 

To further ease the training of LSTM, we propose a variant whose outputs are defined as the residual latent sequences. That is, instead of modeling as in \eqref{eq:sltc}, we define the following generating process:
\begin{align}
\label{eq:latent}
    (\vb_1, \cdots, \vb_T) &= \LSTM\left(\xib_1, \cdots, \xib_T, \yb; \thetab_1\right) \nonumber \\
    \hb_{t+1} &= \hb_{t} + \vb_t~.
\end{align}

\vspace{-0.3cm}
\paragraph{The shared frame-wise decoder}
The second component, GSDC, is a shared frame-wise decoder mapping one latent frame $\hb_t$ to the corresponding action pose $\xb_t$. Specifically, given $\hb_t$ from SLTC, we have, for all $t$,
\begin{align*}
    \tilde{\xb}_t = \Dec(\hb_t, \yb; \thetab_2)~,
\end{align*}
where $\Dec(\cdot, \cdot; \thetab_2)$ represents a decoder implemented as any DNN with parameter $\thetab_2 \subset \thetab$, mapping an input latent frame $\hb_t$ to an output action pose $\tilde{\xb}_t$. In experiments, we use a simple MLP structure for $\Dec$.

\vspace{-0.3cm}
\paragraph{The whole action sequence generator}
Stacking the above two components constitutes our implicit action generator. 
To further enforce smooth transitions, we penalize the generated latent action poses by the changes of consecutive frames, {\it i.e.}, with the following regularizer, similar to \cite{Cai_2018_ECCV}:
{\small\begin{align}\label{eq:reg1}
    \Omega(\{\hb_t\}, \{\tilde{\xb}_t\}) \triangleq \sum_{t=2}^T \left(\sigma_1 \|\hb_t - \hb_{t-1}\|^2 + \sigma_2\|\tilde{\xb}_t - \tilde{\xb}_{t-1}\|^2\right)
\end{align}}
where $\sigma_1$ and $\sigma_2$ control the relative importance of the corresponding regularizer term.

\subsection{Action-Sequence Classifier}
Modeling action-sequence generation and classification simultaneously enables information sharing between the generator and classifier, thus it is expected to be able to boost model performance. As a result, we define a classifier $C$ with a bi-directional LSTM \cite{birnn1997}, whose outputs are further appended with a fully connected layer and a softmax layer for classification. The purpose of adopting the bi-directional LSTM is to effectively model frame-wise relation from two directions, which has been shown more effective than single direction modeling in sequence models \cite{bilstm2015,transbi2014,seqtrans2012}. Please refer to $C_{\phib}$ in Figure~\ref{fig:gan_struc} for a detailed structure of our sequence classifier.

\subsection{Action Discriminator and Model Training}

\paragraph{Bi-directional GAN based training}
The proposed action-sequence generator and classifier constitute a pair of networks that can translate between each other, {\it i.e.}, inverting the classifier achieves the same goal of the generator. To train these two networks effectively, we borrow ideas from bi-directional GAN, and define a discriminator to play adversarial games with the generator and classifier. Specifically, the action-label pairs come from two sources: one starts from a random label and then generates an action sequence via the generator $G_{\thetab}$; the other starts from a randomly-sampled training action sequence and then generates a label via the classifier $C_{\phib}$. Let $q(\yb)$ be a prior distribution over labels\footnote{We adopt a uniform distribution in our method.}; $q_{\thetab}(\tilde{\xb}|\yb)$ be the implicit distribution induced by the generator; $p(\xb)$ be the empirical action-sequence distribution of the training data; and $p_{\phib}(\tilde{\yb}|\xb)$ be the conditional label distribution induced by the classifier given a sequence $\xb$. Our model updates the generator $G_{\thetab}$, the classifier $C_{\phib}$, and the discriminator $D_{\psib}$ alternatively by following the GAN training procedure. Similar to the classifier, the discriminator is also defined by a bidirectional LSTM. The bi-directional GAN is trained to match the joint distributions $q(y)q_{\thetab}(\tilde{\xb}|y)$ and $p(\xb)p_{\phib}(\tilde{\yb}|\xb)$, via the following min-max game:
{\small\begin{align}\label{eq:loss}
    \min_{G_{\thetab}, C_{\phib}}\max_{D_{\psib}}&V(G_{\thetab}, C_{\phib}, D_{\psib}) = \mathbb{E}_{\xb\sim p(\xb), \tilde{\yb} \sim p_{\phib}(\cdot|\xb)}\left[\log D_{\psib}(\xb, \tilde{\yb})\right] \nonumber \\
    &+ \mathbb{E}_{y \sim q(y), \tilde{\xb} \sim q_{\thetab}(\cdot|\yb)}\left[\log(1 - D_{\psib}(\tilde{\xb}, \yb))\right]~.
\end{align}}

\vspace{-0.3cm}
In addition, motivated by CycleGAN \cite{CycleGAN2017} and ALICE \cite{alice2017}, a cycle-consistency loss is introduced:
\vspace{-0.3cm}
\begin{align}\label{eq:reg}
    \mathcal{L}_c \triangleq H(C_{\phib}(G_{\thetab}(\yb, \xib)), \yb)~,
\end{align}
where $H(\cdot,\cdot)$ denotes the cross entropy between two distributions. Combining \eqref{eq:reg1}, \eqref{eq:loss} and \eqref{eq:reg} constitutes the final loss of our model.

\begin{figure}[t]
	\centering
    \begin{minipage}{0.8\linewidth}
        \includegraphics[width=\linewidth]{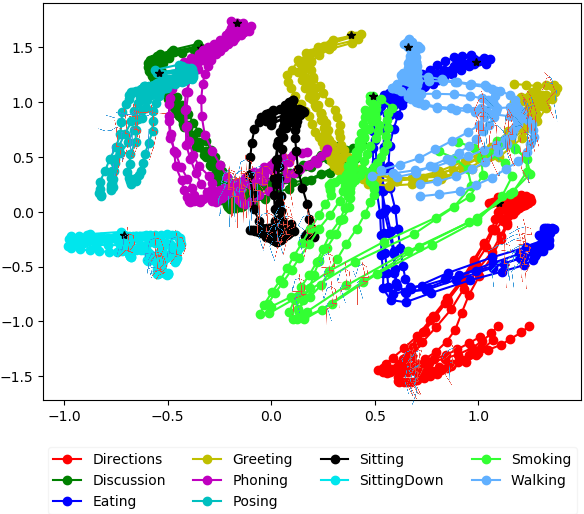}
    \end{minipage}
	
	\caption{Latent space with dimension = 2. The trajectories intercept with each other due to some similar frames in different action sequences.} 
	\label{fig:latdimall}
	\vspace{-0.4cm}
\end{figure}

\begin{figure}
    \centering
    
    \begin{minipage}{0.48\linewidth}
        \includegraphics[width=\linewidth]{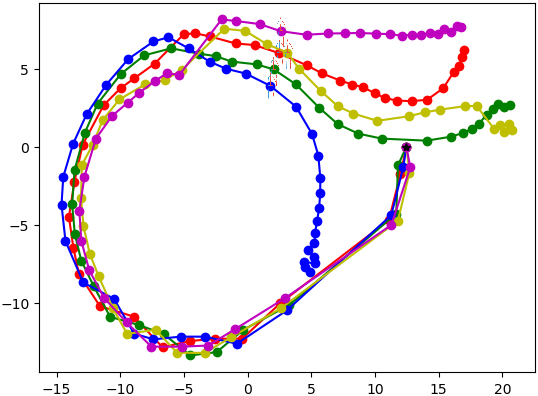}
    \end{minipage}
    \begin{minipage}{0.48\linewidth}
        \includegraphics[width=\linewidth]{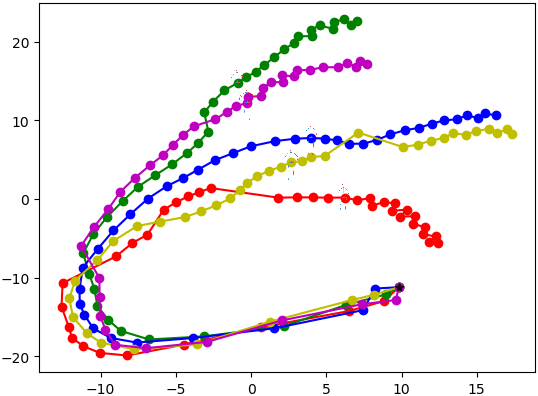}
    \end{minipage}
    \caption{
    Action diversity of generated latent trajectories.  
    Left: Greeting; Right: Posing.}\label{fig:diversity}
    \vspace{-0.4cm}
\end{figure}

\begin{figure}
    \includegraphics[width=0.95\linewidth]{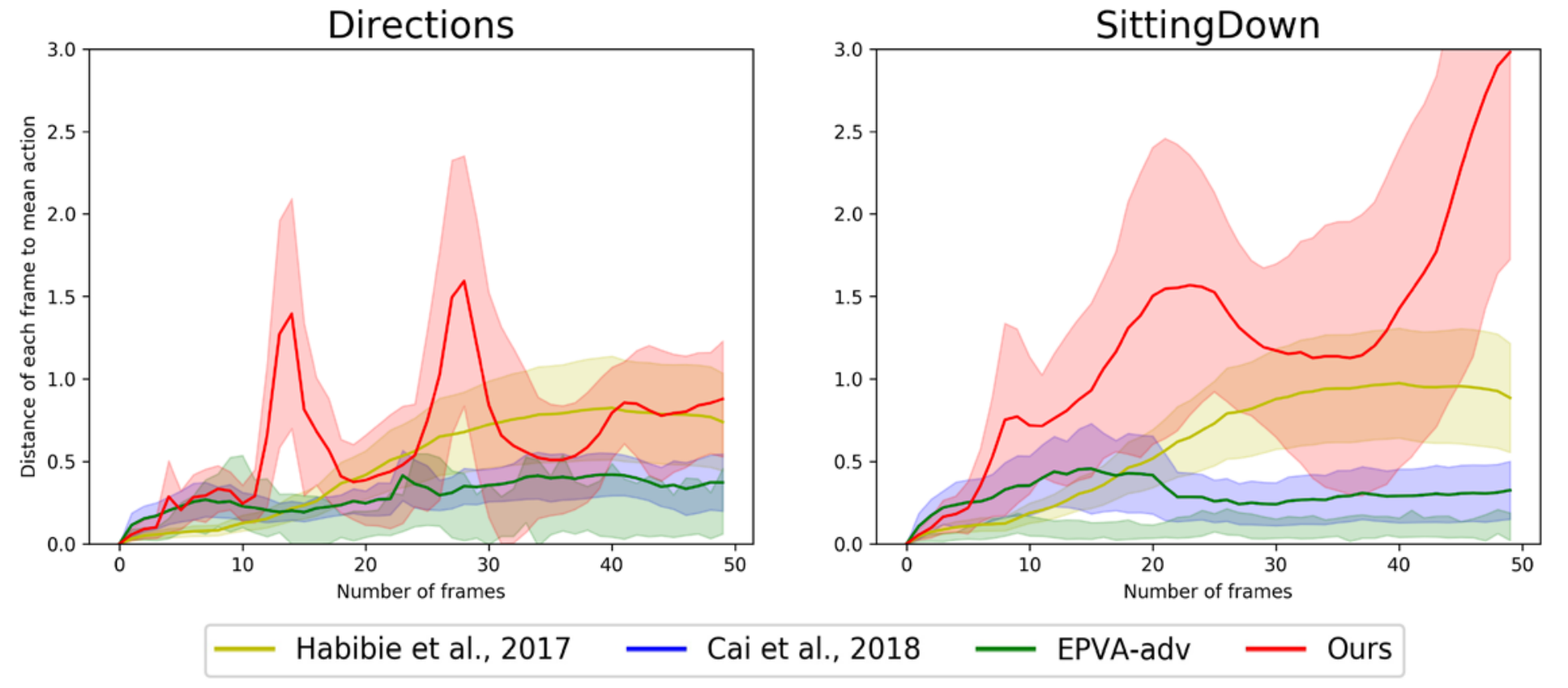}
    \caption{Diversity of generated action sequences.}\label{fig:std_diversity}
    \vspace{-0.3cm}
\end{figure}

\paragraph{Shared frame-wise decoder pretraining}
It is useful to pretrain the shared frame-wise decoder with the training data. To this end, we use the conditioned WGAN-GP model \cite{NIPS2017_wgan} to train a generator to generate independent action poses from a given label. The generator, denoted as $\bar{G}_{\thetab_2}(\cdot)$, corresponds to the shared decoder in our model. To match the input with our frame decoder, we replace the original input $\hb_t$ with a random sample from a simple distribution $p_{h}(\cdot)$, {\it e.g.}, the standard Gaussian distribution. The discriminator, denoted as $\bar{D}(\cdot)$, is an auxiliary network to be discarded after pretraining. The objective function is defined as:
\begin{align}\label{Eq1_wgan}
    &\underset{\bar{G}_{\thetab_2}}{\min} \;\underset{\bar{D}}{\max} \;\bar{V}(\bar{D}, \bar{G}_{\thetab_2}) = \mathbb{E}_{\xb\sim p_{\text{data}}}[\bar{D}(\xb)] -  \\ 
    &\mathbb{E}_{\hb\sim p_h(\hb)}[\bar{D}(\bar{G}_{\thetab_2}(\hb))]  +  \lambda \mathbb{E}_{\xb\sim p_{\text{data}}}[{(\parallel \bigtriangledown_{\xb} \bar{D}(\xb)\parallel}_2 - 1)^2] \nonumber
\end{align}
where $p_{\text{data}}$ denotes the frame distribution of training data; and $\lambda$ controls the magnitude of the gradient penalty to enforce the Lipschitz constraint.

Finally, the whole training procedure of our model is described in Algorithm (See Appendix).

\section{Experiments}\label{exp}

We evaluate our proposed model for diverse action generation on two datasets, in terms of both action-sequence quality and diversity. We also conduct extensive human evaluation for the results generated by different models. Ablation study, implementation details and more results are provided in the Appendix. Code is also made available\footnote{https://github.com/zheshiyige/Learning-Diverse-Stochastic-Human-Action-Generators-by-Learning-Smooth-Latent-Transitions}.



\subsection{Datasets \& Baselines}
\paragraph{Datasets}
We adopt the human-3.6m dataset \cite{Catalin14} and the NTU dataset \cite{Shahroudy_2016_CVPR}. The human-3.6m is a large scale dataset for human activity recognition and analysis. Following \cite{Cai_2018_ECCV}, we subsample the video frames to 16 fps to obtain more significant action variations. 
Our model is trained on 10 classes of actions, including {\em Directions}, {\em Discussion}, {\em Eating}, {\em Greeting}, {\em Phoning}, {\em Posing}, {\em Sitting}, {\em SittingDown}, {\em Walking}, and {\em Smoking}.

The NTU RGB+D is a large action dataset collected with Microsoft Kinect v.2 cameras \cite{Shahroudy_2016_CVPR}. For our purpose, we only use the 2D skeleton locations of 25 major body joints in the corresponding depth/IR frame data. Similar to the human3.6m dataset, we also sample 10 action classes for training and testing, including {\em drinking water}, {\em throw}, {\em sitting down}, {\em wear jacket}, {\em standing up}, {\em hand waving}, {\em kicking something}, {\em jump up}, {\em make a phone call} and {\em cross hands in front}. 
We follow the evaluation protocols of previous literature on this dataset to adopt cross-subject and cross-view recognition accuracy. For the cross-subject evaluation, sequences for training (20 subjects) and testing (20 subjects) come from different subjects. For the cross-view evaluation, the training dataset consists of action sequences collected by two cameras, and the test dataset consists of the remaining data.
After splitting and cleaning missing or incomplete sequences, there are 2260 and 1070 action sequences for training and testing, respectively, for cross-subject evaluation; and there are 2213 and 1117 action sequences for training and testing, respectively, for cross-view evaluation.

\paragraph{Baselines}
Generating action sequences from scratch is a relatively less explored field. The most related models to ours we found are the recently proposed generative models EPVA-adv in \cite{WichersVEL:ICML18}, action generator trained with VAE  \cite{VAEmotion2017} as well as the model proposed in \cite{Cai_2018_ECCV} for the human-action generation. In the experiments, we will compare our model with these three, as well as other specific baselines.

\subsection{Detailed Results}

\begin{figure*}[t]
	\centering
    \includegraphics[width=0.8\linewidth]{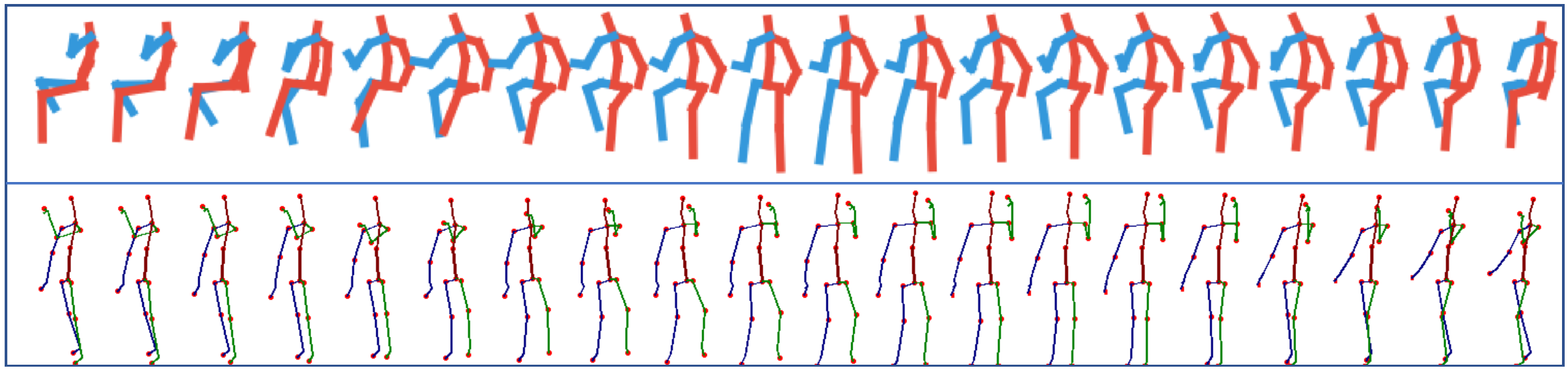}
	\caption{Randomly selected action sequences generated on human3.6 dataset ( First row: Smoking) and NTU RGBD dataset (Last row: Drinking)}.  
	\label{fig:h36_vis}
	\vspace{-0.3cm}
\end{figure*}


\begin{table*}[ht!]
    \centering
    \caption{Comparisons of our model with \cite{VAEmotion2017,Cai_2018_ECCV,WichersVEL:ICML18} in terms of Maximum Mean Discrepancy. (The lower the better.)}
    \vspace{-0.2cm}
    \begin{tabular}{c|c|c|c|c|c|c|c|c|c}
        \hline
        &\multicolumn{3}{c|}{With seed motion}&\multicolumn{6}{c}{Without seed motion}\\
        \cline{2-10}
        &E2E&EPVA&EPVA-adv&E2E&EPVA&EPVA-adv&Habibie et al., 2017&Cai et al., 2018 &Ours\\
        \hline
         $\MMD_{\text{avg}}$&0.304&0.305&0.339&0.991&0.996&0.977&0.452&0.419& {\bf 0.195} \\
         \hline
         $\MMD_{\text{seq}}$& 0.305 &0.326&0.335&0.805&0.806&0.792&0.467&0.436& {\bf 0.218} \\
         \hline
    \end{tabular}
    \label{tab:mmd}
    \vspace{-0.3cm}
\end{table*}

\begin{figure}[ht!]
	\centering
	\subfloat[$p_1$=0.5, $p_2$=0.5]{
	    \label{}
		\includegraphics[width=0.22\textwidth]{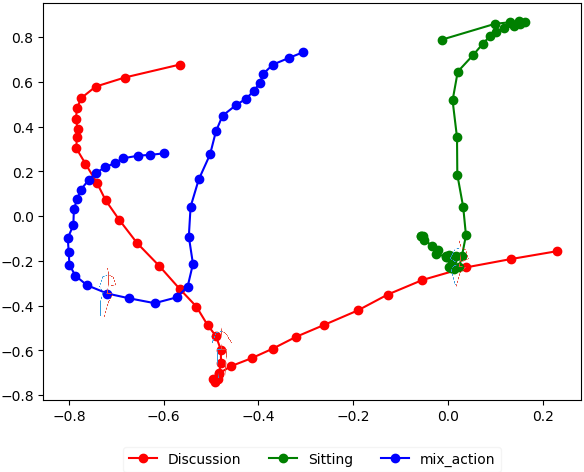}
	}
		\hfill
			\subfloat[$p_1$=0.3, $p_2$=0.7]{
	    \label{}
		\includegraphics[width=0.22\textwidth]{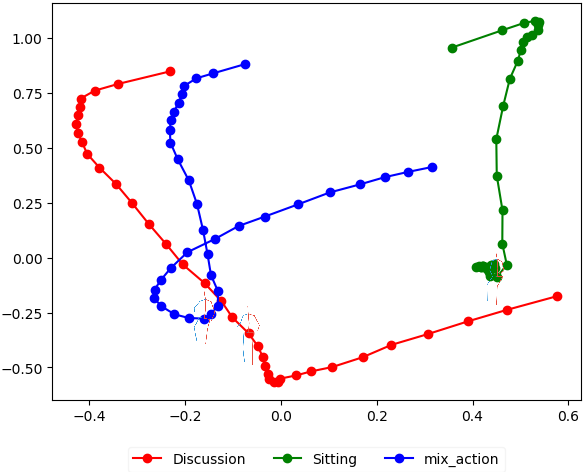}
	}

	\caption{Latent space of mixed classes of actions with different mixing proportions $p_1$ and $p_2$.} 
	\label{fig:lat_mix}
	\vspace{-0.3cm}
\end{figure}

\begin{figure*}[t]
	\centering
	\begin{minipage}{\linewidth}
	    \centering
        \includegraphics[width=0.7\linewidth]{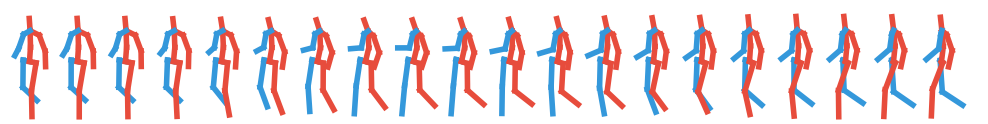}
    \end{minipage}
    \begin{minipage}{\linewidth}
        \centering
        \includegraphics[width=0.7\linewidth]{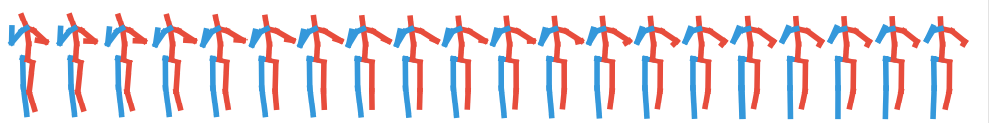}
    \end{minipage}
    \vspace{0.2cm}
    \begin{minipage}{\linewidth}
        \centering
        \includegraphics[width=0.7\linewidth]{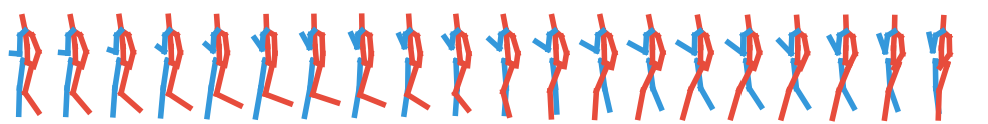}
    \end{minipage}
    \vspace{-0.0cm}
	\caption{Novel mixed action sequences generated on human3.6 dataset. First row: generated sequence of ``Walking''; Second row: generated sequence of ``Phoning''; Third row: generated sequence of mixed ``Walking'' + ``Phoning''.} 
	\label{fig:hu36_mixvis2}
	\vspace{-0.2cm}
\end{figure*}

\paragraph{Latent frame transitions}
To show the effectiveness of our proposed latent-transition mechanism, we visualize the learned latent representations for selected classes.

Latent trajectories of different classes on the Human-3.6 dataset are plotted in Figure \ref{fig:latdimall}, with a latent-frame dimension of 2. More results with higher dimensionalities are provided in the Appendix. It is interesting to observe that for the 2-dimensional-latent-space case, some latent trajectories intercept with each other. This is reasonable because action poses in different action categories might be similar, {\it e.g.}, smoking (green) and eating (blue) in Figure~\ref{fig:latdimall} (left). 

To demonstrate the diversity of the learned action generator, we plot multiple latent trajectories for selected classes in Figure \ref{fig:diversity}, all starting from the same initial point. It is clear that as time goes on, the generated latent frames become more diverse, a distinct property lacking in deterministic generators in most action-prediction models such as \cite{julieta2017motion,pred_eccv2018}.

To better illustrate the diversity of the action sequences, we compare our model with the three recently proposed models \cite{VAEmotion2017,Cai_2018_ECCV,WichersVEL:ICML18}. The mean and variance of each action pose along time for a set of trajectories are plotted in Figure~\ref{fig:std_diversity}. It is remarkable to find that trajectories from our model are much more diverse than the baselines. For a quantitative comparison, the standard derivations for different action classes are listed in the Appendix. More diverse action sequences generation results are provided  in the Appendix. All these results indicate the superiority of our model in generating diverse action sequences.

\vspace{-0.3cm}
\paragraph{Quality of generated action sequences}
We adopt two metrics to measure the quality of the generated actions: action-classification accuracy and the maximum mean discrepancy (MMD) between generated and real action sequences. The latter metric is adapted from measuring the quality of generated images of GAN-based model, and has been used in \cite{pos_iccv2017} for measuring the quality of action sequences. For action classification, we adopt the trained classifier from our model to classify both real (testing) actions and a set of randomly generated actions from our model. A good generator should generate action sequences that endow similar or better classification accuracies than real data. We compare our model with a baseline, which uses the same classifier structure but is trained independently on the training data. For every action class, we randomly sample 100 sequences for testing. The classification accuracies are shown in Table \ref{tab:2 similarity}. It is seen that our model achieves comparable performance on real and generated action sequences, which outperforms the baseline model to a large margin in general.
In some cases, the accuracy on generated action sequence is higher than that of real data is because both the real data and the generated data are involved in the training of classifier under the bidirectional GAN framework \cite{afl2017}.
The cross-subject and cross-view classification accuracies across different classes are 0.824 and 0.885 for our model respectively.

\begin{table*}[ht!]
	\caption{Cross-view and cross-subject  evaluation of classification accuracies on NTU-RGB dataset. Baseline\_T means the independently trained classifier testing on real data, and Baseline\_G means the independently trained classifier testing on generated sequences. Ours\_T means our model testing on real data, and Ours\_G means our model testing on generated sequences.}
	\label{tab:2 similarity}     
	\resizebox{\textwidth}{!}{
	\begin{tabular}{l|l|c|c|c|c|c|c|c|c|c|c}
		
		\hline
	Split	&Data&drinking&throw&sitting down& wear jacket&standing up&hand waving&kick&jump&phoning&cross hands\\
		
		\hline
\multirow{3}*{cross-view}		& Baseline\_T &0.71 &0.90&0.92 & 0.94&0.93&0.73&0.85&0.76&0.87&{\bf 0.92} \\		
	    \cline{2-12}
	    
	    & Baseline\_G& 0.65 & 0.21 & 0.25 & 0.38 & 0.75 & 0.61 & 0.66 & 0.46 & 0.75 & 0.13\\
	    
	    \cline{2-12}
	    
	&	Ours\_T& 0.82& 0.92& 0.94&{\bf 0.96}& 0.95& 0.76& 0.86&{\bf 0.93}& 0.84& 0.88\\
		\cline{2-12}
		&Ours\_G&{\bf 0.87} &{\bf 0.94}&{\bf 0.98}&0.93&{\bf 0.95}&{\bf 0.81}&{\bf 0.97}&0.79&{\bf 0.91}&0.82\\
        \hline\hline
\multirow{3}*{cross-sub}& Baseline\_T &0.76 &0.80&0.90 &0.90&0.92&0.73&0.6&0.67&{\bf 0.86}&0.74 \\		
	    \cline{2-12}
	    & Baseline\_G& 0.60 & 0.15 & 0.12 & 0.57 & 0.74 & 0.67 & 0.34 & 0.26& 0.59 & 0.21 \\
	    \cline{2-12}  
		&Ours\_T  &0.82 &0.65&{\bf 0.93}&{\bf 0.95}& {\bf 0.94}&0.83&0.74&0.69&0.82&{\bf 0.81}\\
		\cline{2-12}
		&Ours\_G& {\bf 0.83}&{\bf 0.86}&0.92& 0.91& 0.90&{\bf 0.87}&{\bf 0.86}&{\bf 0.82}&0.81&0.76\\

		\hline
		
	\end{tabular}
	}
\end{table*}

The MMD measures the discrepancy of two distributions based on their samples (the generated and real action sequences in our case). Since our data are represented as sequences, we proposed two sequence-level MMD metrics (more details in Appendix \ref{appendix: mmd}). Following \cite{pos_iccv2017}, we vary the bandwidth from $10^{-4}$ to $10^9$ and report the maximum value computed. We compared our model with \cite{VAEmotion2017,Cai_2018_ECCV,WichersVEL:ICML18}. \cite{WichersVEL:ICML18} contains three models: E2E, EPVA and EPVA-adv. These models require a few seed frames as inputs to the generator. To adapt these model to our setting, we define two variants: 1) given labels and only the first frame as seed input to their models; 2) given only labels but no frames as seed input. The results are reported in Table~\ref{tab:mmd}. It is interesting to see that without the need of a seed action-pose, our model obtains much lower MMD scores than the baselines with no seed action-poses. Figure \ref{fig:h36_vis} further plots some examples of generated action sequences on the two datasets, which further demonstrates the high quality of our generated actions.
\vspace{-0.3cm}
\paragraph{Novel action generation by mixing action classes}
Another distinct feature of our model is its ability to generate novel action sequences by specifying the input class variable $\yb$. One interesting way for this is to mix several action classes such that the elements satisfy $\yb_k \geq 0$ and $\sum_k\yb_k = 1$. When feeding such a soft-mixed label into our model, due to the smoothness of the learned latent space, one would expect the generated sequence contains all features from the mixing classes. To demonstrate this, we consider mixing two classes. Figure~\ref{fig:lat_mix} plots the latent trajectories for different mixing coefficients. As expected, the trajectories of mixing classes smoothly interpolate between the original trajectories. For better visualization, let the first two classes correspond to {\em walking} and {\em phoning}. We set $\yb \triangleq (0.5, 0.5, 0, \cdots, 0)$ and generate the corresponding action sequences by feeding it into the generator. The generated sequences are shown in Figure \ref{fig:hu36_mixvis2}. It is interesting to see that the sequence with mixing classes indeed contains actions with both hands and legs, which correspond to walking and phoning, respectively.

\begin{wraptable}{r}{5.5cm}
\caption{Human evaluations.}
\scalebox{0.9}{
\begin{tabular}{ccc}\\ \toprule  
Model & Average Score\\ \midrule
\cite{VAEmotion2017}&2.445 \\ 
\cite{WichersVEL:ICML18}&2.387 \\  
\cite{Cai_2018_ECCV}&2.847 \\  
Ours & \textbf{3.378}\\
\bottomrule
\end{tabular}
}

\vspace{-4mm}
\label{table:human evaluation}

\end{wraptable}

\paragraph{Human evaluation} 
We run perceptual studies on Amazon Mechanical Turk (AMT) to assess the realism of generated actions. There were 120 participants in this test for three-round evaluations. Every worker was assigned some collection of evaluation tasks, each of which consists of four videos generated by one of the four models, including \cite{VAEmotion2017,WichersVEL:ICML18,Cai_2018_ECCV} and ours. The worker was asked to evaluate each group of videos with a scale from 1 to 5. The higher the score, the more realistic of an action. Table \ref{table:human evaluation} summarizes the results, which clearly shows the superiority of our method over others. Scoring standard and detailed experiment design are provided in the Appendix.

\paragraph{Ablation study}
We conduct extensive ablation study to better understand each component of our model, including smoothness term, cycle consistency loss, and residual latent sequence prediction. More details are described in the Appendix.

\section{Conclusion}
We propose a new framework for stochastic diverse action generation, which induces flexible implicit distributions on generated action sequences. Different from existing action-prediction methods, our model does not require conditional action poses in the generation process, although it can be easily generalized to this setting. Within the core is a latent-action generator that learns smooth latent transitions, which are then fed to a shared decoder to generate final action sequences. Our model is formulated within the bi-directional GAN framework, which contains a sequence classifier that simultaneously learns action classification. Experiments are conducted on two accessible datasets, demonstrating the effectiveness of the proposed model, and obtaining better results compared to related baseline models.

{\small
\bibliography{egbib}
\bibliographystyle{aaai}}

\newpage

\appendix

\begin{table*}
	\caption{Standard derivation of generated action sequences}
	\label{tab:variance}     
	\setlength{\tabcolsep}{1mm}
	\begin{tabular}{c|c|c|c|c|c|c|c|c|c|c}
		\hline
		
		Data&Directions&Discussion&Eating & Greeting &Phoning& Posing&Sitting&SittingDown&Smoking&Walking\\
		\hline
		
		\cite{VAEmotion2017} & 9.10&13.18&12.44&7.12&11.33&8.11&9.38&10.45&9.47&8.50\\		
	    \hline
	    \cite{Cai_2018_ECCV} & 5.26&13.57&13.26&11.19&10.49&19.37&9.93&12.58&8.00&8.21\\		
	    \hline
	    EPVA-adv \cite{WichersVEL:ICML18}& 7.42&3.77&18.98&9.37 &2.22&1.87&2.48&2.83&5.01&5.93\\
	    \hline
		Ours & \textbf{17.74} &\textbf{21.70}&\textbf{30.45}&\textbf{67.92}&\textbf{20.53}&\textbf{22.73}&\textbf{14.08}&\textbf{33.62}&\textbf{16.50}&\textbf{12.09}\\
		\hline
	\end{tabular}
    \label{tab:variance}
\end{table*}

\begin{table*}[ht!]
    \centering
    \caption{Ablation study of the smoothness term in our model in terms of Maximum Mean Discrepancy. The lower the better.}
    \begin{tabular}{c|c|c|c|c}
        \hline
        &\multicolumn{4}{c}{Smoothness term ablation study}\\
        \cline{2-5}
        &no smoothness&only action &only latent &latent and action\\
        \hline
         $\MMD_{\text{avg}}$&0.798 &0.396 &0.214  &\bf{0.195} \\
         \hline
         $\MMD_{\text{seq}}$&1.254 &0.566 &0.259& \bf{0.218}\\
         \hline
    \end{tabular}
    \label{tab:ablation_smooth}
\end{table*}

\begin{table*}[ht!]
    \centering
    \caption{Ablation study of residual latent transition prediction in terms of Maximum Mean Discrepancy. The lower the better.}
    \begin{tabular}{c|c|c}
        \hline
        &\multicolumn{2}{c}{residual latent prediction ablation study}\\
        \cline{2-3}
        &direct latent &residual latent\\
        \hline
         $\MMD_{\text{avg}}$&0.975 &\bf{0.195}\\
         \hline
         $\MMD_{\text{seq}}$&0.829 &\bf{0.218}\\
         \hline
    \end{tabular}
    \label{tab:ablation_res}
\end{table*}


\section{Algorithm}\label{appendix: algorithm}

\begin{algorithm}[ht!]
	\caption{Stochastic action generation via learning smooth latent transitions.}\label{transitive}
	\begin{algorithmic}
		\REQUIRE Generator $G_{\thetab}$; Discriminator $D_{\psib}$; classifier $C_{\phib}$. Training data $\{(\xb^{(i)}, \yb^{(i)})\}$. Number of updating steps $k$ for discriminator; the weight $\sigma_1$ and $\sigma_2$ for the regularization loss; and the weight $\gamma$ for the cycle consistency loss.
		\STATE Pretrain the shared frame-wise decoder.
        \FOR{$t = 0$ to $T$}
            \FOR{$k = 0$ to $K$}
             \STATE Sample minibatch of $m$ noise samples $\{\xib^{1},\cdots,\xib^{m}\}$.
             \vspace{-4mm}
             \STATE Sample minibatch of $m$ action sequences $\{(\xb^{(1)}, \yb^{(1)}),\cdots,(\xb^{(m)}, \yb^{(m)})\}$ from the training data.
             \STATE Update the discriminator by stochastic gradient ascent:
             \vspace{-0.3cm}
             {\small\begin{align*}
                 \hspace{-0.4cm}\nabla_{\psib} \frac{1}{m} \sum\limits_{i=1}^{m}[\log D_{\psib}(\xb^{(i)}, \yb^{(i)})+ \log(1-D_{\psib}(G_{\thetab}(\xib^{i}, \yb^{(i)})))]
            \end{align*}}
            \ENDFOR
            \STATE Sample minibatch of $m$ noise samples $\{\xib^{1},\cdots,\xib^{m}\}$.
            \STATE Update the generator and classifier parameters by stochastic gradient descent: 
            {\small\begin{align*} 
            \hspace{-0.3cm}\nabla_{\thetab, \phib} &\frac{1}{m} \sum\limits_{i=1}^{m}\left[\log(1-D_{\psib}(G_{\thetab}(\xib^{i}, y^{(i)})))+
             \Omega(\{\hb_t^{(i)}\}, \{\xb_t^{(i)}\})\right. \\
            & \left. + \gamma\left(H(C_{\phib}(\xb^{(i)}), \yb^{(i)})+  H(C_{\phib}(G_{\thetab}(\xib^{i}, \yb^{(i)})), \yb^{(i)})\right)\right]
            \end{align*}} 
        \ENDFOR
	\end{algorithmic}
\end{algorithm}

\section{Maximum Mean Discrepancy (MMD)}\label{appendix: mmd}

\begin{align*}
    \MMD_{\text{avg}} &\triangleq \dfrac{1}{SK} \Sigma_{j=1}^K\Sigma_{i=1}^S \MMD_u \left[ \mathcal{F}, X_i^j, Y_i^j\right] \\
    \MMD_{\text{seq}} &\triangleq \dfrac{1}{K} \Sigma_{j=1}^K \MMD_u \left[ \mathcal{F}, X^j_{\text{seq}}, Y^j_{\text{seq}}\right]~,
\end{align*}
where $X^j_{\text{seq}}$ and $Y^j_{\text{seq}}$ denote generated and real test sequences of action class $j$, respectively; $X^j_i$ and $Y^j_i$ denote the $i^{th}$ frame of the generated and real test sequences of action class $j$, respectively; and $\MMD_u \left[ \mathcal{F}, X, Y\right]$ is defined as an unbiased MMD estimator \cite{DBLP:journals/jmlr/GrettonBRSS12}:
\vspace{-0.2cm}
\begin{align*}
    \MMD_u^2 \left[ \mathcal{F}, X, Y\right]\triangleq \dfrac{1}{m(m-1)}\Sigma_{i=1}^m\Sigma_{j\neq i}^m k(x_i, x_j)
    \\+ \dfrac{1}{n(n-1)}\Sigma_{i=1}^n\Sigma_{j\neq i}^n k(y_i, y_j)
    - \dfrac{2}{mn}\Sigma_{i=1}^m\Sigma_{j=1}^n k(x_i, y_j)~
\end{align*}
with $k( \cdot, \cdot)$ the Gaussian kernel.

\section{Implementation Details}\label{appendix: implementation details}

We use Adam optimization algorithm \cite{adam2015} for learning the whole network parameters. The weight of gradient penalty $\lambda$ for the WGAN-GP is set to be 10. The learning rate for optimizing the generator and discriminator loss of the WGAN-GP is set to be 0.001. The learning rate is set to be 0.0001 for optimizing the generator and discriminator loss of our sequence generation model. The weight $\gamma$ for the cycle consistency loss is set to be 0.1 and the weight $\sigma_1$ and $\sigma_2$ for regularization loss are set to be 0.05 and 0.00005 respectively. The number of hidden units for the fully connected layers is set to 1024 for the LSTM discriminator and classifier. The number of hidden units for the LSTM generator is set to be 256.

\section{More Experiment Results}\label{appendix: more experiment results}
\begin{figure*}[t]
	\centering
    	\subfloat[Direction]{
	    \label{}
		\includegraphics[width=1\textwidth]{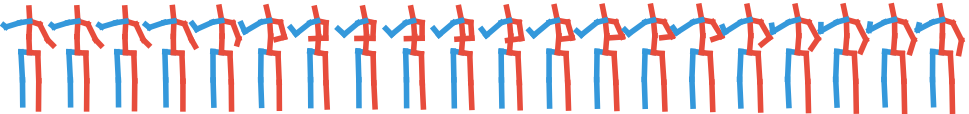}
	}
		\hfill
    
	\subfloat[Greeting]{
	    \label{}
		\includegraphics[width=1\textwidth]{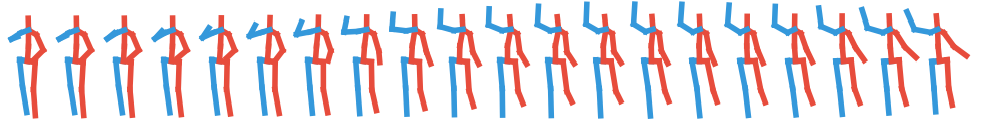}
	}
		\hfill
			\subfloat[Sitting]{
	    \label{}
		\includegraphics[width=1\textwidth]{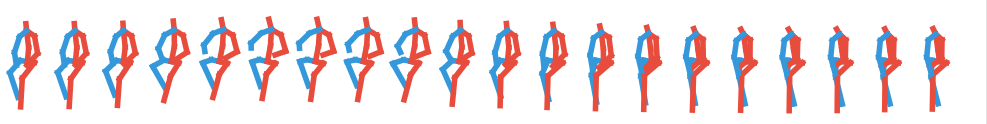}
	}
	\hfill
		\subfloat[Eating]{
	    \label{}
		\includegraphics[width=1\textwidth]{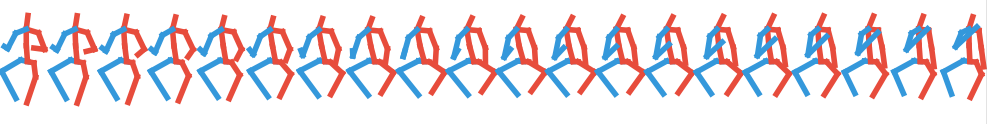}
	
	}
	\caption{Several action sequences generated by training on human3.6 dataset} 
	\label{fig:h36_vis_1}
\end{figure*}

\begin{figure*}[t]
	\centering
	\subfloat[Hand waving]{
	    \label{}
		\includegraphics[width=1\textwidth]{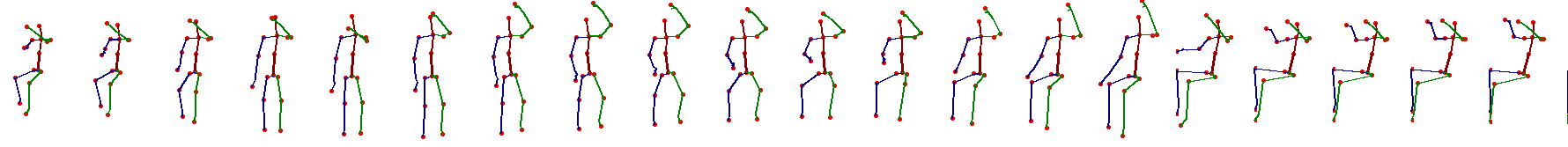}
	}
		\hfill
			\subfloat[throw]{
	    \label{}
		\includegraphics[width=1\textwidth]{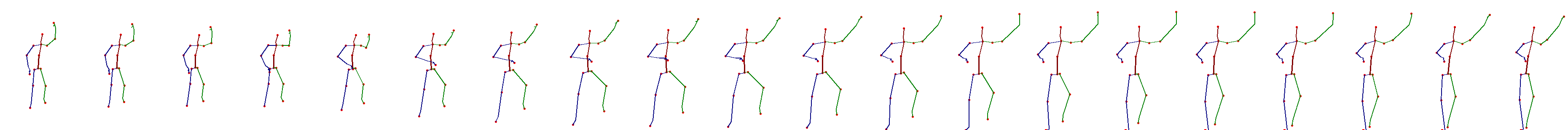}
	}
	\hfill
		\subfloat[Sit down]{
	    \label{}
		\includegraphics[width=1\textwidth]{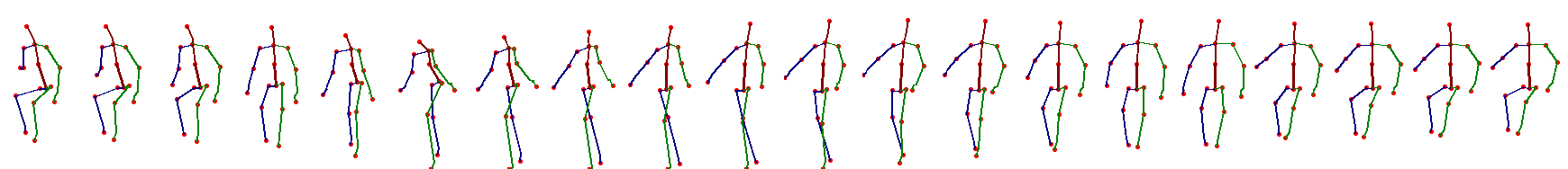}
	
	}
	\caption{Several action sequences generated by training on NTU RGBD dataset} 
	\label{fig:ntu_vis1_1}
\end{figure*}

\begin{figure*}
    \centering
    \subfloat[walking]{\includegraphics[width=0.23\textwidth]{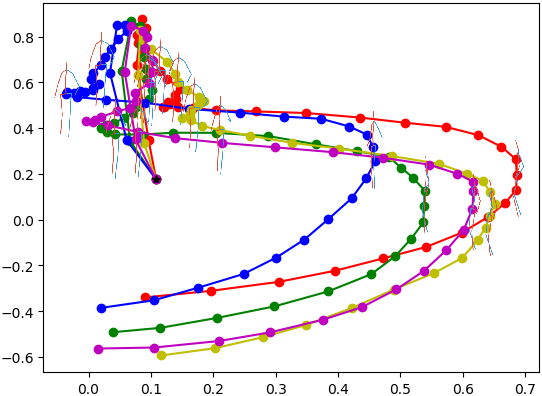}}
    \hfill
    \subfloat[smoking]{\includegraphics[width=0.23\textwidth]{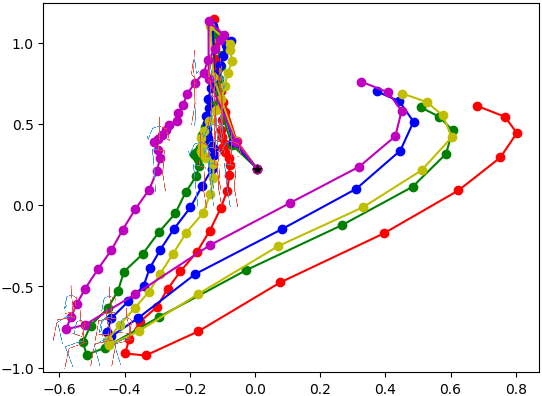}}
    \hfill
    \subfloat[greeting]{\includegraphics[width=0.23\textwidth]{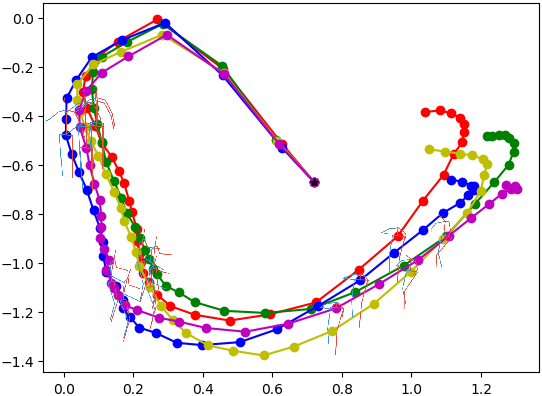}}
    \hfill
    \subfloat[posing]{\includegraphics[width=0.23\textwidth]{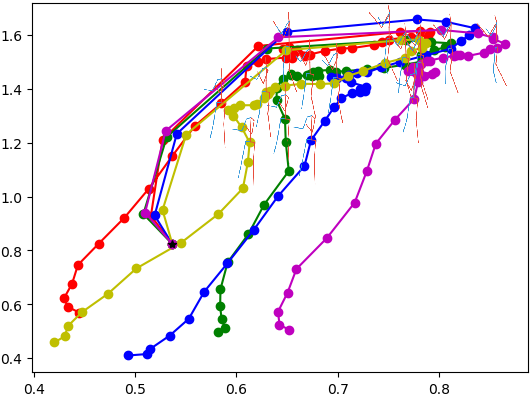}}
    \vfill
    \subfloat[walking]{\includegraphics[width=0.23\textwidth]{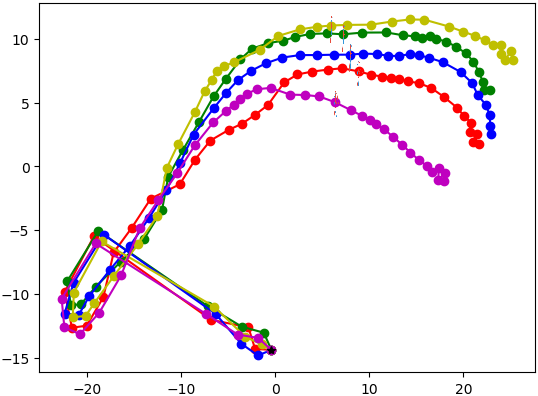}}
    \hfill
      \subfloat[smoking]{\includegraphics[width=0.23\textwidth]{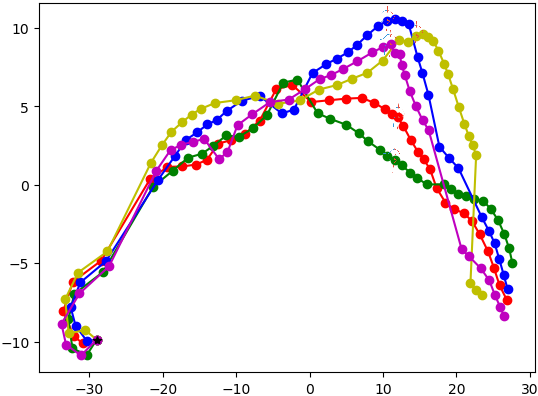}}
    \hfill
     \subfloat[greeting]{\includegraphics[width=0.23\textwidth]{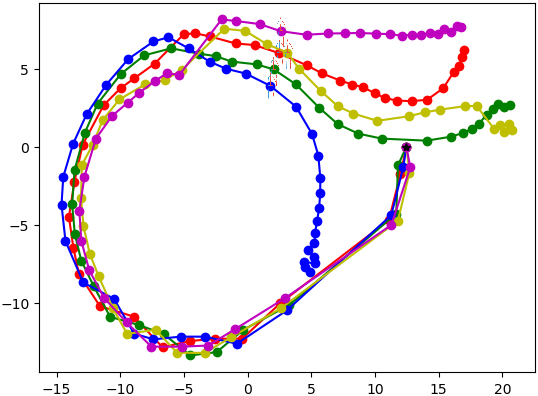}}
    \hfill
     \subfloat[posing]{\includegraphics[width=0.23\textwidth]{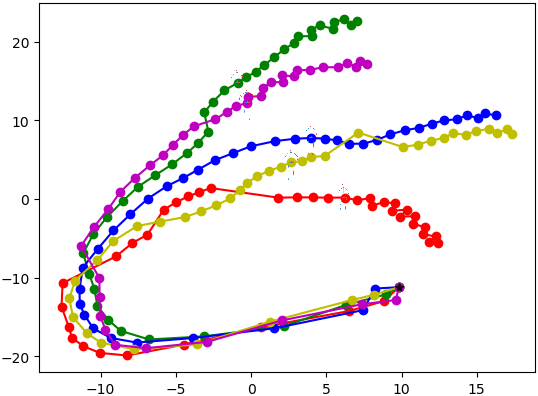}}
    \vfill
    \subfloat[walking]{\includegraphics[width=0.23\textwidth]{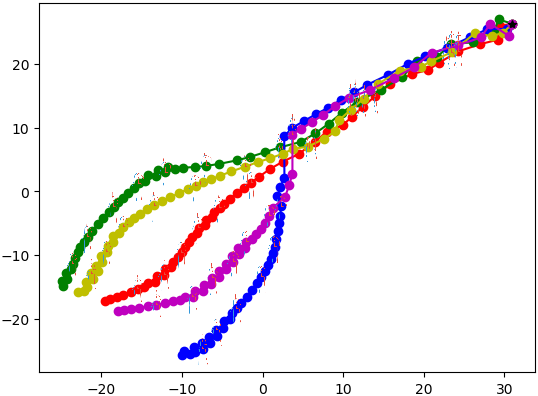}}
    \hfill
      \subfloat[smoking]{\includegraphics[width=0.23\textwidth]{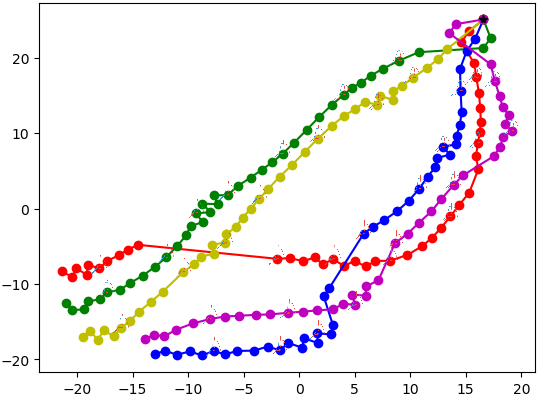}}
    \hfill
     \subfloat[greeting]{\includegraphics[width=0.23\textwidth]{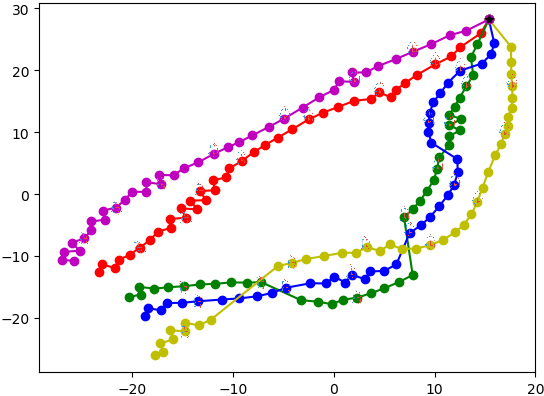}}
    \hfill
     \subfloat[posing]{\includegraphics[width=0.23\textwidth]{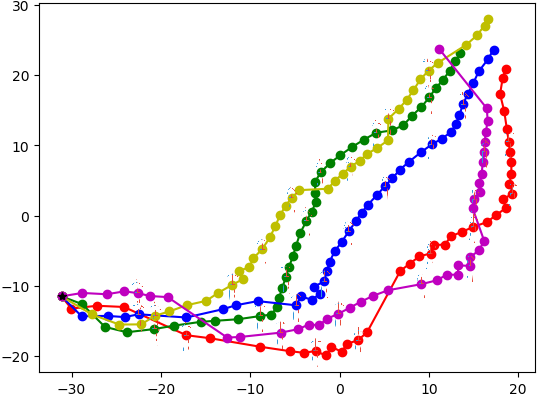}}
    \hfill
    \caption{\label{fig:diversity_1}
    Action diversity of generated latent space with different latent dimensions. The latent space dimension of first row, second row, third row is 2, 6 and 12 respectively.}
    \vspace{-0.2cm}
\end{figure*}

\begin{figure*}[t]
	\centering
	\begin{minipage}{\linewidth}
        \includegraphics[width=1.0\linewidth]{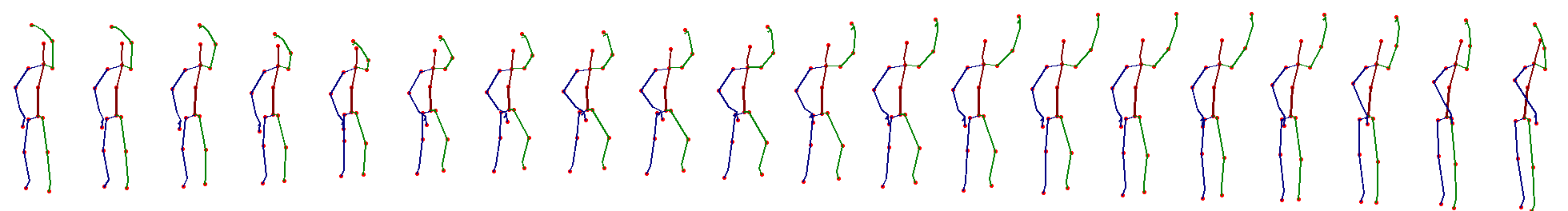}
    \end{minipage}
    \begin{minipage}{\linewidth}
        \includegraphics[width=1.0\linewidth]{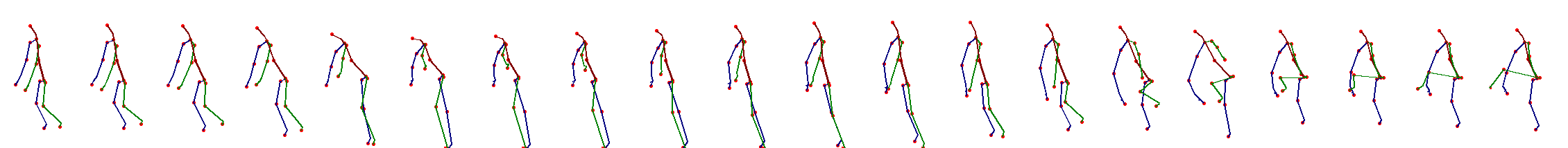}
    \end{minipage}
    \vspace{0.2cm}
    \begin{minipage}{\linewidth}
        \includegraphics[width=1.0\linewidth]{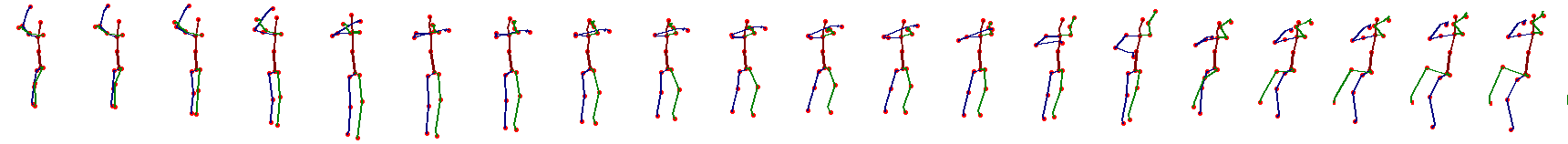}
    \end{minipage}
	\caption{Novel mixed action sequences generated on NTU RGBD dataset. First row: generated sequence of ``Throw''; Second row: generated sequence of ``Kick''; Third row: generated sequence of mixed ``Throw'' + ``Kick''.} 
	\label{fig:ntu_mixvis2}
	\vspace{-0.3cm}
\end{figure*}

The novel actions mixed with the two actions "Throw" and "Kick" trained on NTU RGB+D dataset are shown in Figure \ref{fig:ntu_mixvis2}. It is also interesting to see that the sequence with mixing classes indeed contains actions with both hands and legs, which correspond to "Throw" and "Kick", respectively. Diverse generated latent space with different latent dimensions are shown in Figure \ref{fig:diversity_1}. Figure \ref{fig:h36_vis_1} and \ref{fig:ntu_vis1_1} shows other examples of generated action sequences on the two datasets. Table \ref{tab:variance} shows the standard deviation of the distance of the generated action sequences to the mean action for each action class.

\begin{figure*}[t]
	\centering
    	\subfloat[Eating sequence 1]{
	    \label{}
		\includegraphics[width=1\textwidth]{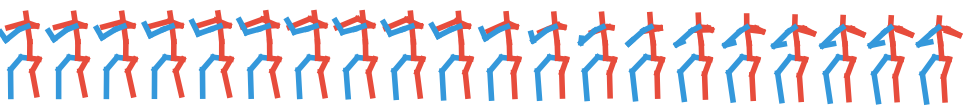}
	}
		\hfill
    
	\subfloat[Eating sequence 2]{
	    \label{}
		\includegraphics[width=1\textwidth]{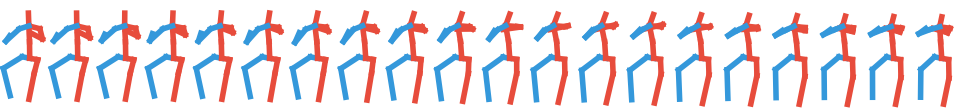}
	}
		\hfill
			\subfloat[Eating sequence 3]{
	    \label{}
		\includegraphics[width=1\textwidth]{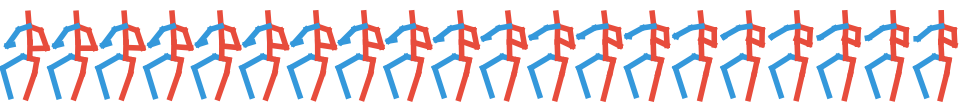}
	}
	\hfill
		\subfloat[Eating sequence 4]{
	    \label{}
		\includegraphics[width=1\textwidth]{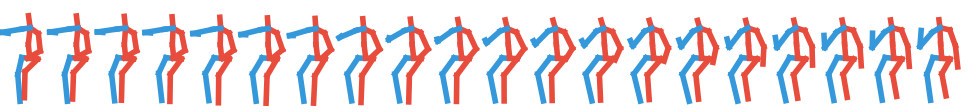}
	
	}
	
		\hfill
		\subfloat[Eating sequence 5]{
	    \label{}
		\includegraphics[width=1\textwidth]{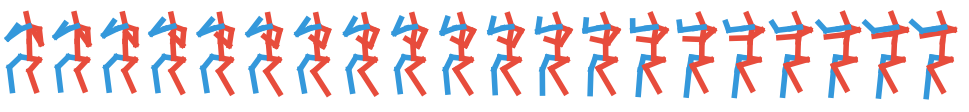}
	
	}
	
	\caption{Diverse action sequences generated by training on human3.6 dataset} 
	\label{fig:h36_diverse_eating}
\end{figure*}

\begin{figure*}[t]
	\centering
    	\subfloat[Smoking sequence 1]{
	    \label{}
		\includegraphics[width=1\textwidth]{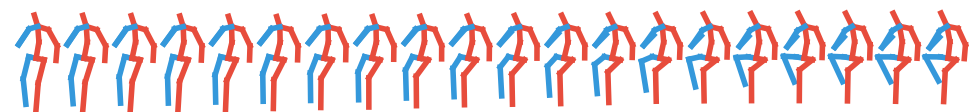}
	}
		\hfill
    
	\subfloat[Smoking sequence 2]{
	    \label{}
		\includegraphics[width=1\textwidth]{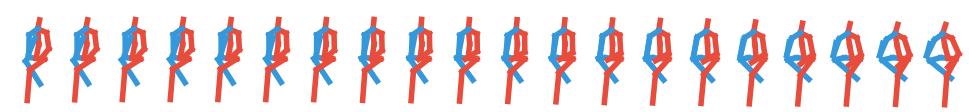}
	}
		\hfill
			\subfloat[Smoking sequence 3]{
	    \label{}
		\includegraphics[width=1\textwidth]{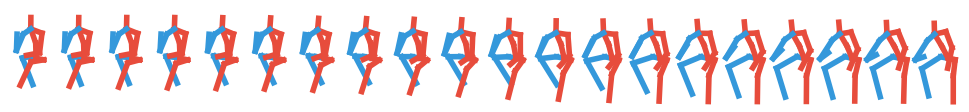}
	}
	\hfill
		\subfloat[Smoking sequence 4]{
	    \label{}
		\includegraphics[width=1\textwidth]{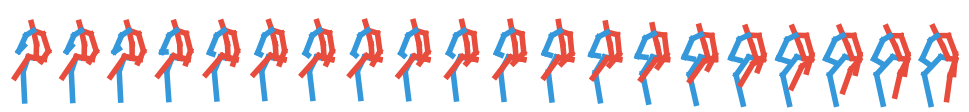}
	
	}
	
		\hfill
		\subfloat[Smoking sequence 5]{
	    \label{}
		\includegraphics[width=1\textwidth]{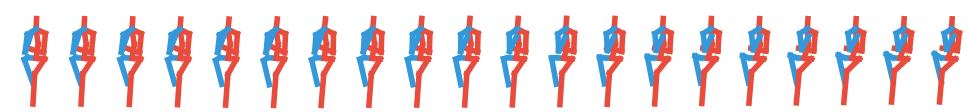}
	
	}
	
	\caption{Diverse action sequences generated by training on human3.6 dataset} 
	\label{fig:h36_diverse_smoking}
\end{figure*}

\begin{figure*}[t]
	\centering
    	\subfloat[Direction sequence 1]{
	    \label{}
		\includegraphics[width=1\textwidth]{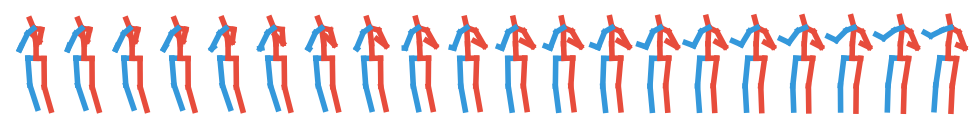}
	}
		\hfill
    
	\subfloat[Direction sequence 2]{
	    \label{}
		\includegraphics[width=1\textwidth]{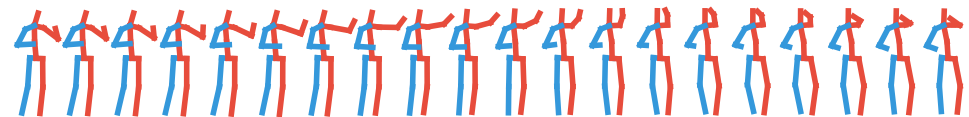}
	}
		\hfill
			\subfloat[Direction sequence 3]{
	    \label{}
		\includegraphics[width=1\textwidth]{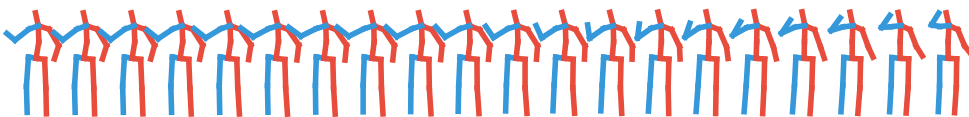}
	}
	\hfill
		\subfloat[Direction sequence 4]{
	    \label{}
		\includegraphics[width=1\textwidth]{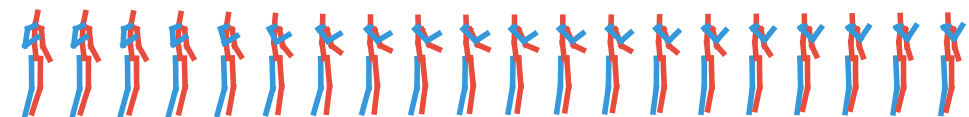}
	
	}
	
		\hfill
		\subfloat[Direction sequence 5]{
	    \label{}
		\includegraphics[width=1\textwidth]{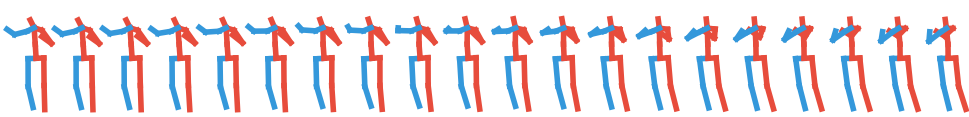}
	
	}
	
	\caption{Diverse action sequences generated by training on human3.6 dataset} 
	\label{fig:h36_diverse_direction}
\end{figure*}

Figure \ref{fig:h36_diverse_eating}, \ref{fig:h36_diverse_smoking}, \ref{fig:h36_diverse_direction} shows more diverse action generation results on human3.6 dataset.

\begin{figure*}[t]
	\centering

	\begin{minipage}{0.46\linewidth}
        \includegraphics[width=\textwidth]{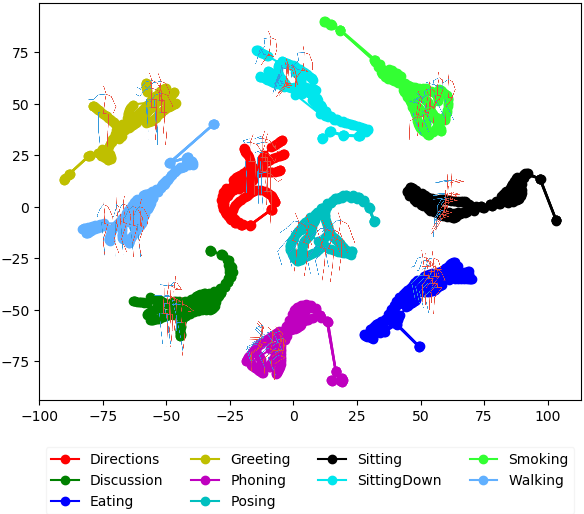}
    \end{minipage}
    \begin{minipage}{0.46\linewidth}
        \includegraphics[width=\textwidth]{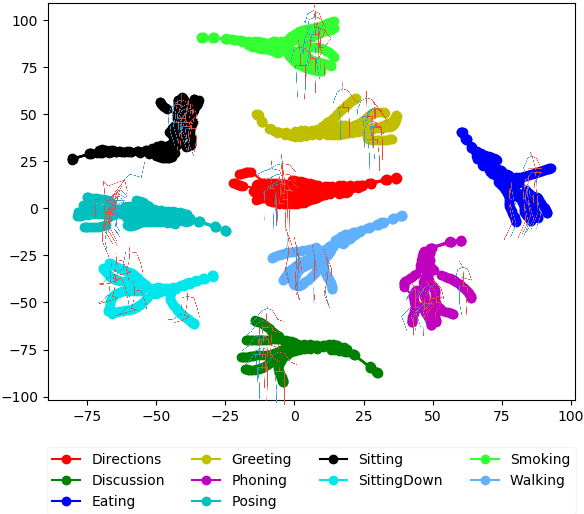}
    \end{minipage}
	\caption{Latent space with different dimensions. Left: dim = 6; Right: dim = 12.} 
	\label{fig:latdimall_appendix}
\end{figure*}

For latent dimensions larger than 2, the latent trajectories from different classes are found to be separated quite well. Figure \ref{fig:latdimall_appendix} shows more visualization results of latent space in higher dimension. Apart from the impact of tSNE, we suspect in a higher latent space, our model is flexible enough to separate trajectories of different classes, as the input of the generator contains label information.


\section{Ablation Study}\label{appendix: ablation study}

We run ablation experiments over the components of our model to better understand the effects of each component.

\paragraph{Effect of smooth regularizer:} We remove the regularizer for consecutive frames. Table \ref{tab:ablation_smooth} shows the ablation study for smoothness term in the loss function. "no smoothness" means that we remove both the regularization term on the latent and action space. "only action" means that we only add regularizer on the action space. "only latent" means that we only add regularizer on the latent space. "latent and action" means that we add regularizer on both action and latent space.

The results show that the regularization of both latent space and real action space is better than only regularize the latent or real action space or no regularization at all.

\paragraph{Effect of cycle consistency loss:} We add or remove the classifier in the model, i.e., add or remove the cycle consistency loss. Table \ref{tab:2 similarity} shows the results of the ablation study for the cycle consistency loss, which is shown in Equation \eqref{eq:reg}.

\paragraph{Effect of transition residual:} We study the comparison of predicting the latent transition residual and predicting the latent transition directly. Table \ref{tab:ablation_res} shows the results of the ablation study for latent transition residual. "direct latent" means that we use LSTM to predict the latent transition directly. "residual latent" means that we use LSTM to predict the residual of latent transition as in Equation \eqref {eq:latent}. 

\section{Human Evaluation}\label{appendix: human evaluation}
\begin{figure*}[h!]
    \centering
    \includegraphics[width=1\textwidth]{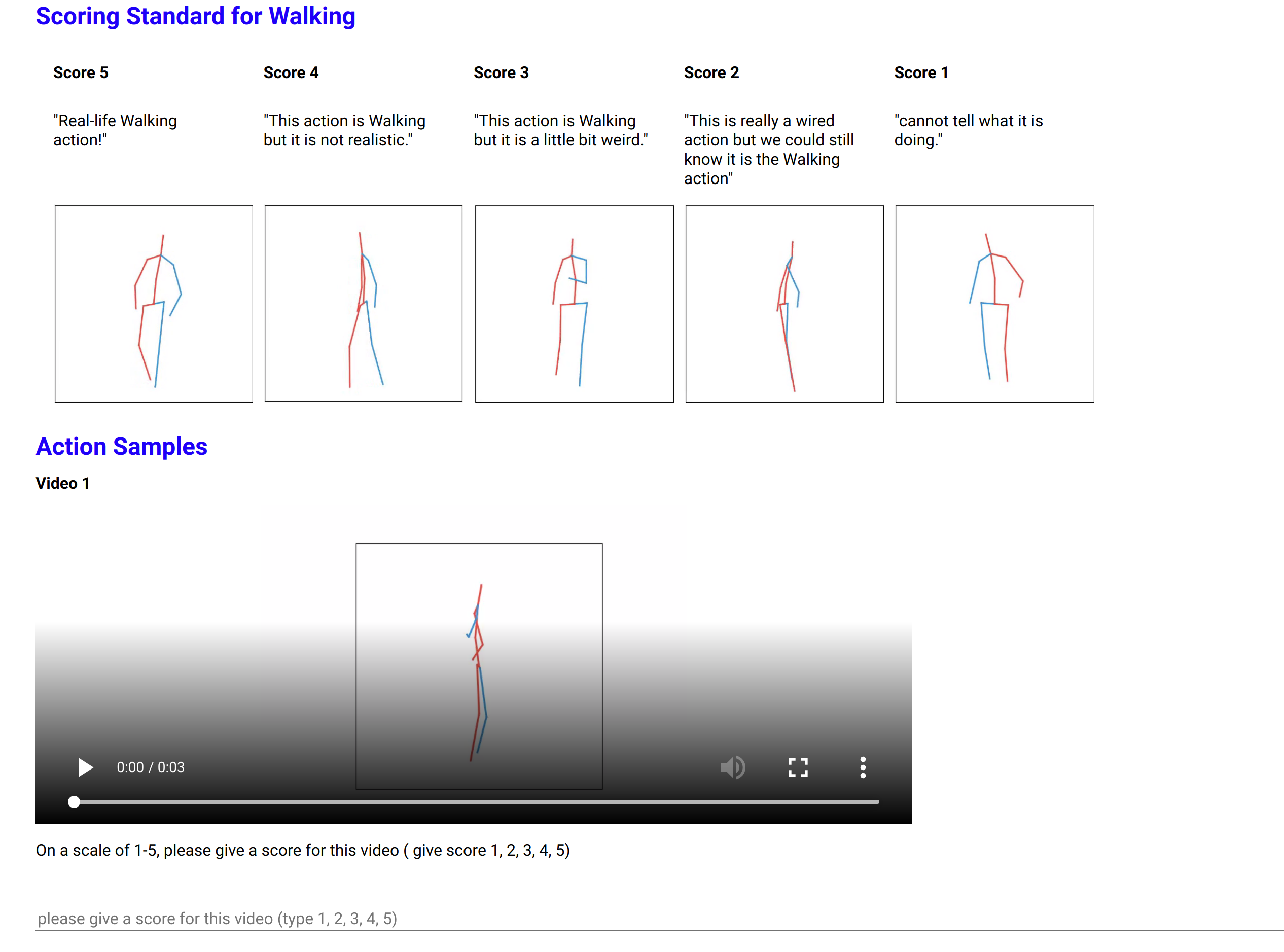}
    \caption{Human evaluation screenshot for Amazon Mechanical Turk.}
    \label{fig:human evaluation}
\end{figure*}

Figure~\ref{fig:human evaluation} shows experiment design for Human Evaluation. At each page, we give five scoring standard with word descriptions as well as video examples to make sure workers share the same grading standard. After that, we provide four sample videos from each model for the blind test.

\newpage

\end{document}